\documentclass{article}

     \usepackage[preprint]{neurips_2019}

\usepackage[utf8]{inputenc} 
\usepackage[T1]{fontenc}    
\usepackage{hyperref}       
\usepackage{url}            
\usepackage{booktabs}       
\usepackage{amsfonts}       
\usepackage{nicefrac}       
\usepackage{microtype}      
\usepackage{todonotes}
\usepackage{latexsym,bm,amsmath, amssymb, amsthm, bbm, epsf, graphicx, color,hhline}
\usepackage{tabularx}
\usepackage{import}

\newcommand{\be}{\begin{equation}}
\newcommand{\ee}{\end{equation}}
\newcommand{\ba}{\begin{align}}
\newcommand{\ea}{\end{align}}
\newcommand{\bea}{\begin{eqnarray}}
\newcommand{\eea}{\end{eqnarray}}

\newcommand{\vb}{{\boldsymbol{b}}}

\newcommand{\vu}{{\boldsymbol{u}}}
\newcommand{\vv}{{\boldsymbol{v}}}

\newcommand{\vQ}{{\boldsymbol{Q}}}

\newcommand{\vtheta}{{\boldsymbol{\theta}}}

\newcommand{\vpi}{{\boldsymbol{\pi}}}

\newcommand{\vphi}{{\boldsymbol{\phi}}}

\newcommand{\vGamma}{{\boldsymbol{\Gamma}}}

\usepackage{natbib}
\setcitestyle{numbers}
\setcitestyle{square}

\title{Inverse Rational Control:\\ Inferring What You Think from How You Forage}

\author{
  Zhengwei Wu\\ 
  Baylor College of Medicine, Rice University\\
  Houston, TX 77030 \\
  \texttt{Zhengwei.Wu@bcm.edu} \\
  \And
 Paul Schrater\\
  University of Minnesota\\
  Minneapolis, MN 55455\\
  \texttt{schrater@umn.edu} 
  \AND
   Xaq Pitkow\\ 
  Baylor College of Medicine, Rice University\\
  Houston, TX 77030 \\
  \texttt{xaq@rice.edu} 
}

\begin{document}
\maketitle

%

%

\begin{abstract}
Complex behaviors are often driven by an internal model, which integrates sensory information over time and facilitates long-term planning. Inferring an agent's internal model is a crucial ingredient in social interactions (theory of mind), for imitation learning, and for interpreting neural activities of behaving agents. Here we describe a generic method to model an agent's behavior under an environment with uncertainty, and infer the agent's internal model, reward function, and dynamic beliefs. We apply our method to a simulated agent performing a naturalistic foraging task. We assume the agent behaves rationally --- that is, they take actions that optimize their subjective utility according to their understanding of the task and its relevant causal variables. We model this rational solution as a Partially Observable Markov Decision Process (POMDP) where the agent may make wrong assumptions about the task parameters. Given the agent's sensory observations and actions, we learn its internal model and reward function by maximum likelihood estimation over a set of task-relevant parameters. The Markov property of the POMDP enables us to characterize the transition probabilities between internal belief states and iteratively estimate the agent's policy using a constrained Expectation-Maximization (EM) algorithm. We validate our method on simulated agents performing suboptimally on a foraging task currently used in many neuroscience experiments, and successfully recover their internal model and reward function. Our work lays a critical foundation to discover how the brain represents and computes with dynamic beliefs.
\end{abstract}

\section{Introduction}
In an uncertain and partially observable environment, animals learn to plan and act based on limited sensory information. To better understand these natural behaviors and interpret their neural mechanisms, it would be beneficial to estimate the internal model and reward function that explains animals' behavioral strategies. In this paper, we use Partially Observed Markov Decision Processes (POMDP) to model animals as rational agents acting optimally under possibly incorrect assumptions about the world. We then solve an inverse POMDP problem to infer these internal assumptions and estimate the dynamics of internal beliefs. We call this approach (model-based) Inverse Rational Control (IRC), because we infer the reasons or motivation that explains an agent's suboptimal behavior.

We use an agent's actions to learn its internal model for the world. This model includes its assumed stochastic dynamics of the partially observable world variables and the agent's subjective assessment of rewards and costs. Other past efforts have addressed reduced versions of this problem, either solving parts of the problem like learning dynamics {\it or} subjective rewards, or assuming a fully observable environment. Our work solves all of these problems together.


At the highest level, by positing a rational but possibly mistaken agent, our approach is closest to a Bayesian Theory of Mind (BToM) \cite{daunizeau2010observing, huszar2010mind,baker2011bayesian,baker2017rational}. Previous work assumed static latent variables that were unknown until fully observed \cite{baker2017rational}; we allow for dynamic latent variable and partial observability. Some of this earlier work used simpler trial-structured tasks like perceptual decision-making \cite{daunizeau2010observing, huszar2010mind}, whereas our method can infer models that agents use to make long-term plans and choose sequences of actions. Our approach learns both stochastic dynamics and subjective reward functions simultaneously, whereas prior work in BToM learned only subjective rewards \cite{baker2017rational}. Other work \cite{rafferty2015inferring} inferred beliefs, whereas we infer both dynamic beliefs and the internal model that gave rise to those dynamics. Finally, \cite{baker2017rational} used a brute force grid search to find the best fitting model, whereas we provide a general probabilistic prescription for learning a POMDP agent model using Expectation Maximization, including an analytic expression for the objective function gradient.

Other well-known inverse problems address parts of our Inverse POMDP. Inverse Reinforcement Learning (IRL) tackles the problem of learning how an agent judges rewards and costs based on observed actions \cite{russell1998learning}, but assumes a known dynamics model \cite{choi2011inverse, babes2011apprenticeship}. Conversely, Inverse Optimal Control (IOC) learns the agent's internal model for the world dynamics \cite{dvijotham2010inverse} and observations \cite{schmitt2017see}, but assumes the reward functions. In \cite{herman2016inverse, reddy2018you} both reward function and dynamics were learned, but only the fully-observed MDP case is explored, whereas we solve the more difficult partially-observed case.

To solve our Inverse POMDP problem, we cast it as a maximum-likelihood optimization where the reward functions and latent stochastic dynamics can be learned with gradient descent methods \cite{babes2011apprenticeship} to identify the parameters that best explain the animal behaviors under a known model-based task structure. We use the Expectation-Maximization (EM) algorithm \cite{dempster1977maximum}, specifically a modified version of the Baum-Welch algorithm, to estimate the parameters of the internal model, and infer the posterior of the latent states.

Our method should serve as a valuable tool for creating interpretable models for real agents performing tasks with natural features. To demonstrate our approach, we apply it to an ecologically relevant foraging task that requires sensitivity to past rewards, current observations, and an internal memory state. 


In Section \ref{sec:BehavioralModeling}, we define the task structure as a general POMDP, and represent an agent's solution of this task using a Belief MDP (a Markov Decision Process over belief states). The method to infer the internal model is explained in Section \ref{sec:InversePOMDP}. We next applied our method to a naturalistic foraging task that we define in Section \ref{sec:ForagingTask}. Results of these numerical experiments are shown in Section \ref{sec:Experiments}, followed by a brief discussion.

\section{Behavioral modeling}
\label{sec:BehavioralModeling}
There is a one-to-one correspondence between a POMDP and an MDP operating on the space of beliefs (a Belief MDP). By using this equivalence, we are able to define belief states and their dynamics, and compute the rational policy by which an artificial agent chooses actions, given its reward function and action costs. 

\subsection{Modeling behavior as a POMDP}

In a POMDP, since the world is not fully observed, the agent must create an internal representation of latent states in the world. Identifying the content of such an internal representation will help to identity how these task-relevant variables are encoded in neural responses. Instead of solving directly for the POMDP, we use the mapping between POMDP and a Belief Markov Decision Processes (Belief MDP) to solve for the policies that describe the best actions for the animal's internal model.

In a POMDP in discrete time, the state of the world, $s$, follows dynamics described by transition probability $T(s',s,a)=P(s'|s, a)$, where $s'$ is the new state, $s$ is the current state, and $a$ is the action selected by an agent. However, the agent does not have direct access to the world state $s$, but must infer it from measurements $o$. The sensory information for the agent depends on the action and the world state following probability distribution $O(o,s,a)=P(o|s, a)$. Upon taking action $a$, the agent receives an immediate reward $r = R(s, a)$. The goal of an agent solving a POMDP is to choose actions that maximize the long-term expected reward $E[\sum_{t = 0}^{\infty} \gamma^t r_t]$ based on a temporal discount factor $0<\gamma<1$. 


\subsection{Belief MDP}
In a partially observable environment, an agent can only act on the basis of past actions and observations. The concept of {\it belief}, which is a posterior distribution over world states $s$ given sensory information, concisely summarizes the information that can be used by agents during decision making. If there are $|S|$ possible values, then the beliefs states take on continuous values in an $|S| -1$ dimensional simplex.  Mathematically, we write the belief $\vb(s)$ as a vector with length equal to the number of states. The $i$-th element of the belief vector $\vb$ is the probability that the state at time $t$ is $s = i$ given the past sensory information,
\begin{align}
b_t^{i}(s) = P(s_t = i|o_{1:t}).
\end{align} 
For simplicity, we will use $b$ instead of $\vb(s)$ to denote the belief state at a certain time in the rest of the paper when there is no ambiguity. The belief state representation allows a POMDP problem to be mapped onto an MDP problem with fully observable states. Here the state is now a belief state instead of true world state, and the process is known as a {\it Belief MDP}. In a belief MDP, The policy $\pi(a|b)$ describes the probability of choosing an action under a certain belief state $b$.  We use the ``state-action value'', $Q_{\pi}(b, a)$, to quantify how much total future reward can be obtained by taking action $a$ from belief state $b$ and then following a particular policy $\pi$ subsequently. This value function under the optimal policy $\pi^*$ can be expressed in a recursive form using the Bellman equation \cite{bellman1957dynamic}:
\begin{align} \label{eq:BellmanEquationMax}
Q_{\pi^*}(b, a) = \sum\limits_{s} R(s, a)b + \gamma \sum\limits_{o}P(o|a, b)\max_{a'}Q_{\pi^*}(b', a')
\end{align}
where $\gamma$ is the temporal discount factor, and $b'$ is the next belief state after observing $o$ and taking action $a$, defined as 
\begin{align}
\label{eq:beliefupdate}
{b'_{a}}^{o} = \dfrac{P(o|s', a)\sum_{s}P(s'|s, a)\,b}{\sum_{s'}P(o|s', a)\sum_{s}P(s'|s, a\,b}.
\end{align}

When there is no ambiguity, the dependence of the belief state on $a$ and $o$ will be suppressed. In a Belief MDP, the belief state is a probability distribution and thus takes on continuous values. By discretizing the belief space, we will be to solve the belief MDP problem with standard MDP algorithms \cite{bellman1957dynamic, howard1964dynamic}.

\section{Inferring an agent's internal model and preferences}
\label{sec:InversePOMDP}

The dynamics of the belief states and the policy are dependent on a set of parameters $\vtheta$. We assume that the agent knows the structure of the task, but not the parameters. 
Inferring the agent's parameters $\vtheta$ enables us to better understand the internal model and the reward funciton, and further infer the latent belief of the agent. This can be viewed as a maximum likelihood estimation problem. Due to the Markov property of the belief MDP model, this estimation problem can be analyzed using a hidden Markov model (HMM) where the belief state is a latent variable. 

\subsection{EM algorithm for IRC}


The EM algorithm \cite{dempster1977maximum} enables us to solve for the parameters that give best explanation of the observed data, while inferring unobserved states in the model. Denote by $l(\vtheta)$ the likelihood of the observed data, where $\vtheta$ are the parameters of the model which include both assumptions about the world dynamics and the parameters determining the sizes of rewards and action costs. We alternately update the parameters $\vtheta$ that improve the expected complete-data log-likelihood and the posterior over latent states based on the estimated parameter. 
\begin{figure*}[h]
   \centering
   \includegraphics[width=5.5in]{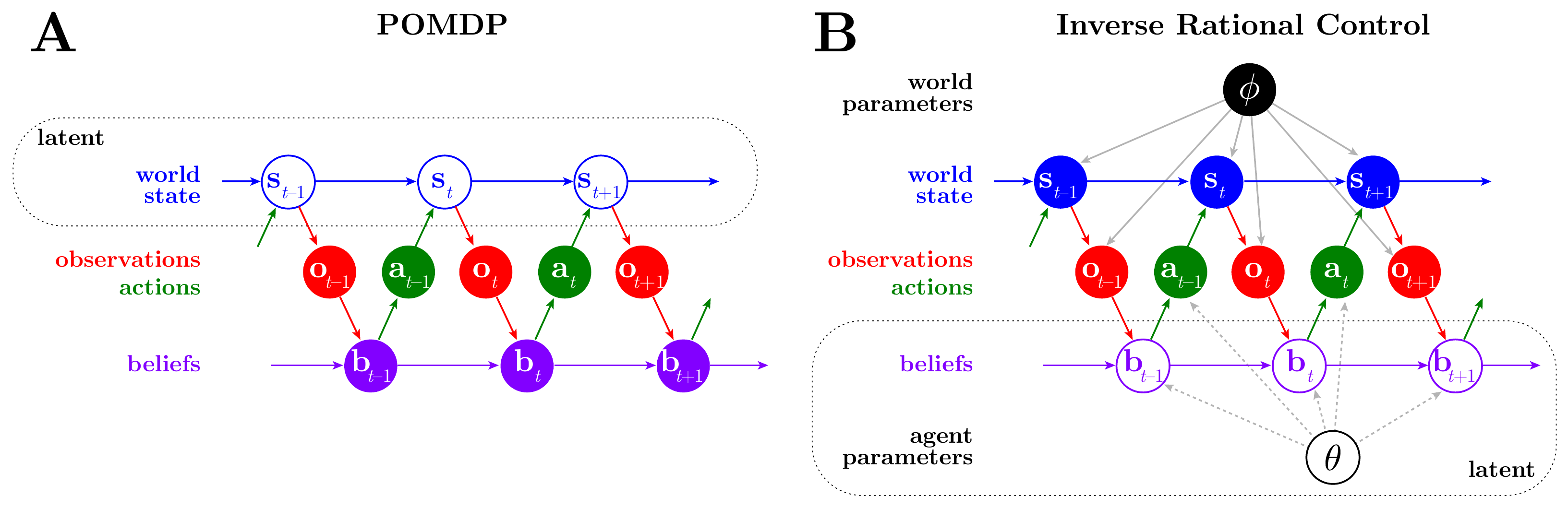}
   \caption{Graphical model of a POMDP ({\bf A}) and the Inverse Rational Control problem ({\bf B}). Empty circles denote latent variables, and solid circles denote observable variables. The real world dynamics depends on parameter $\vphi$, while that of the agent depends on parameter $\vtheta$.}
   \label{fig:trellis}
\end{figure*}

Figure \ref{fig:trellis} depicts a graphical model of the relevant variables. Let $b$ be the belief state, which is the latent variable in our belief MDP model, and let $a$ and $o$ be the action and sensory information. Faced with a POMDP, the agent only gets partial observations about the world state $s$ through its sensory observations $o$ (Figure \ref{fig:trellis}A). The experimenter, however, can observe the agent's actions and sensory observations (Figure \ref{fig:trellis}B). This corresponds to assuming that the agent's uncertainty is dominated by external sensory noise, not by the sensory noise we cannot measure.

According to the EM algorithm, in the E-step the estimated parameters $\vtheta^{\text{old}}$ from the previous iteration determine the posterior distribution of the latent variable given the observed data $P(b_{1:T}|a_{1:T}, o_{1:T}, \vtheta^{\rm{old}})$, where the sub-index $1:t$ indicates the time range of the samples. In the M-step, the observed data log-likelihood function to be maximized reduces to
\begin{align}
l(\vtheta) = \mathcal{Q}(\vtheta, \vtheta^{\text{old}}) + H(P(b_{1:T} |a_{1:T}, o_{1:T}, \vtheta^{\rm{old}})).
\end{align}
To be consistent with \cite{robert2014machine}, we use $\mathcal{Q}(\vtheta, \vtheta^{\text{old}})$ as the auxiliary function that describes the expected complete data likelihood, and $H(\cdot)$ is the entropy of the posterior of the latent variable.\footnote{Unfortunately, the conventional notations in EM and reinforcement learning collide here, both using the same letter: this $\mathcal{Q}(\vtheta, \vtheta^{old})$ auxiliary function is different from the $Q$-value function in the MDP model, and is denoted in the Calligraphic font.}

With fixed parameters $\vtheta^{\text{old}}$ from the previous iteration, the entropy of the latent state $H(P(b_{1:T} |a_{1:T}, o_{1:T},\vtheta^{\text{old}})$ is fixed. As a result, we need to update parameter $\vtheta$ that maximizes function $\mathcal{Q}(\vtheta, \vtheta^{\text{old}})$ in the M-step. During iterations of EM algorithm, the value of the log-likelihood $l(\vtheta)$ always increases to a (possibly local) maximum.

For the Belief MDP, the expected complete data log likelihood $\mathcal{Q}(\vtheta, \vtheta^{\text{old}})$ can be decomposed into transition probabilities and policies at each time due to the Markov property. The $\mathcal{Q}$-auxiliary function can therefore be expressed as:
\begin{align}\label{Qauxiliary}
\mathcal{Q}(\vtheta,\vtheta^{\rm old})&=
\langle \log P(b_{1:T},a_{1:T}, o_{1:T},s_{1:T}|\vtheta,\vphi) \rangle_{P(b_{1:T}|a_{1:T}, o_{1:T},s_{1:T},\vtheta^{\rm old},\vphi^{\rm old})}
\end{align}
where $\phi$ are the parameters in the experiment setup that determine the world dynamics. 
Note here for consistency, we include $\vphi^{\rm old}$ in the sub-index term that indicate the posterior of beliefs given the observations. However, since $\vphi$ are fixed in the experiment and known in the analysis, they are constant across iterations. Ultimately, these experimental parameters do not affect our inference of the agent's model when we condition on the observations and actions.

We can factorize the log probability using the graphical model structure,
\begin{align}
\mathcal{Q}(\vtheta,\vtheta^{\rm old})=
\Big\langle &\sum\nolimits_t \log P(b_{1}, o_{1},s_{1}|\vtheta,\vphi)\\
+&\sum\nolimits_t \log P(a_t|b_t,\vtheta)\label{eq:likelihood_policy} \\ 
+&\sum\nolimits_t \log P(b_{t+1}|b_t,a_{t+1},o_t,\vtheta)\label{eq:likelihood_tran} \\ 
+&\sum\nolimits_t \log P(o_{t+1}|s_t,\vphi)\\
+&\sum\nolimits_t \log P(s_{t+1}|s_t,a_t,\vphi)\Big\rangle_{P(b_{1:T}| a_{1:T}, o_{1:T},s_{1:T},\vtheta^{\rm old},\vphi^{\rm old})}
\end{align}

In the $\mathcal{Q}$ auxiliary function, the term in  (\ref{eq:likelihood_tran}) depends only on the parameters describing the Bayesian update for the state dynamics and the agent's uncertainty about it, while the policy terms in (\ref{eq:likelihood_policy}) depend on both the dynamic parameters and reward functions. These terms depend on the parameters $\vtheta$ implicitly. Instead of solving for the optimal $\vtheta$ in a losed form, we use gradient descent to update the parameter $\vtheta$ in the M-step. 

\subsection{Derivatives of policy}
When the policy is optimal, the term $P(a_t|b_t,\vtheta)$ is a delta function, so the derivative of the policy does not exist. As a result, we approximate the optimal policy using a softmax or Boltzmann policy with a small learnable temperature $\tau$.  The softmax introduces an additional sub-optimality of the agent: instead of choosing the action that brings the maximal expected reward, the agent has some chance of choosing a worse action. Under the softmax policy $\pi_{\rm sfm}$, the actions taken from belief state $b$ follow the distribution 
\begin{align} \label{softmax}
\pi_{\text{sfm}}(a| b) = P(a|b ) \sim \dfrac{e^{Q_{\pi_{\text{sfm}}}(b, a)/ \tau}}{\sum\nolimits_{a'}e^{Q_{\pi_{\text{sfm}}}(b,  a')/ \tau}}.
\end{align}

If we can calculate the derivative of the $Q$-value function with respect to the parameter set $\vtheta$, we are able to get the derivatives of the policy. Similarly to the Bellman equation (\ref{eq:BellmanEquationMax}) based on the optimal policy, the $Q$-value function under a softmax policy can also be expressed in a recursive way, replacing the max with an average.

Denote the vectorized version of $Q(b, a)$ and $\pi(a| b)$ as $\vQ^V$ and $\vpi^V$. Differentiating with respect to $\theta_i\in \vtheta$ on both sides of the Bellman equation gives us:
\begin{align}\label{ValueDerivative}
\nonumber \dfrac{\partial \vQ^V}{\partial \theta_i}
=
\mathbf{c}_i^V
+ \gamma
& \vGamma(P(o|b, a))
\Big(\text{Diag}(\vQ^V)
\dfrac{\partial\vpi^V}{\partial \vQ^V} + \text{Diag}(\vpi^V)\Big)
\dfrac{\partial \vQ^V}{\partial \theta_i} 
\end{align}
where $\mathbf{c}_i^V$ is a vectorized version of the matrix $\mathbf{c}_i(b,a)$ with
\begin{align}
\mathbf{c}_i(b,a)=\sum\limits_{s} \dfrac{\partial R(s, a)}{\partial \theta_i}b + \gamma \sum\limits_{o}\dfrac{\partial P(o|a, b)}{\partial \theta_i}\sum\limits_{a'}\pi_{\text{sfm}}(a'| b')Q_{\pi_{\rm{sfm}}}(b', a')
\end{align}
and $\vGamma(P(o|b, a))$ is a function containing repeated blocks of the observation model $P(o|b, a)$. The detailed derivation is given in the Supplementary Material. 


By reorganizing equation (\ref{ValueDerivative}), we can see that the derivative of the $Q$-value function with respect to the parameters can be solved analytically as a linear function of the known quantities. Using the chain rule, the gradient of the policy can be obtained in this way. We then use this gradient in the M step of the EM algorithm to estimate the internal model parameters.

\section{Application to foraging}
\label{sec:ForagingTask}
\begin{figure}[h]
   \centering
   \includegraphics[width=5in]{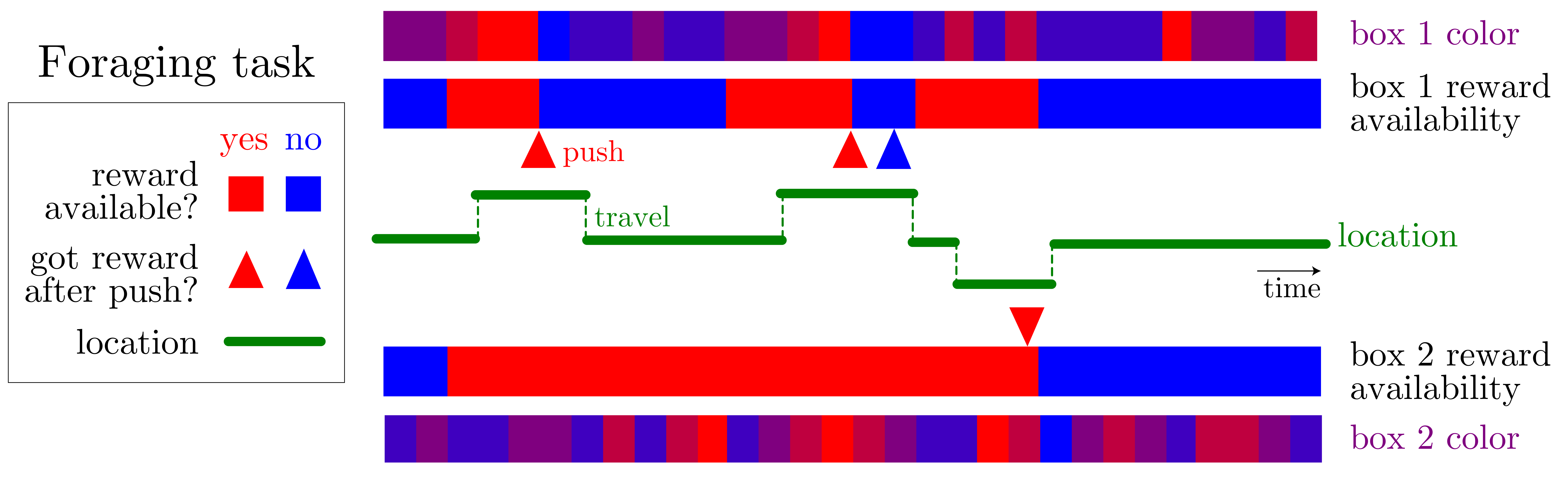}
   \vspace{-5pt}
   \caption{An example of task data. The reward availability in each of two boxes evolves according to a telegraph process, switching between available (red) and unavailable (cyan), and colors give the animal an ambiguous sensory cue about the reward availability. The agent may travel between the locations of the two boxes.  When the button is pressed to open a box, the agent receives any available reward.}
   \label{fig:taskData}
\end{figure}

We applied our method to the specific setting of a task in which an animal can forage at either of two locations (`feeding boxes') which may have hidden food rewards that appear and disappear with a certain rate, based on ambiguous sensory cues about the reward availability (Figure \ref{fig:taskData}). 

To define the Belief MDP for this `two-box' task, we need to define the belief states, actions and rewards. The states must represent the agent's location, whether it has obtained food from the boxes, and also whether food is available in each box. Since the agent knows its location exactly, and whether it has obtained food, we only need an uncertain belief representation for the unobserved food availability in each box.

We assume there are three possible locations for the agent: the locations of boxes 1 and 2, and a middle location 0. 
We include a small `grooming' reward for staying at the middle location to allow the agent to stop and rest.

A few discrete actions are available to the animal: it can push a button to open a box to either get reward or observe its absence, it can move toward a new location, or it can do nothing (Figure \ref{fig:taskData}). Traveling and pushing a button to open the box each have an associated cost. This disincentivizes the agent from rapidly repeating actions that might access reward.

In addition to the cost of actions, there are several parameters that are related to the experiment setting. The food availability in each box follows a telegraph process, alternating between available and unavailable at uniform switching rates $\gamma$ and $\epsilon$. The sensory cue, color, displayed on each box follows a binomial distribution conditioned on the true state of the food availability. The mean and variance is determined by the true state and the distributions under the two states should have enough overlap so that the animal cannot depend only on the color cue to anticipate the food availability. Two more parameters are associated with the color model. Details about the observation model can be found in the Supplementary Information. We assume the agent knows this task structure, but may mistakenly assume different values for the parameters of the task setup. We know neither its internal model for those parameters not its subjective costs associated with each action.



For computational tractability, we discretize beliefs in each box into $N$ states. This is sensible also computationally, since it is unlikely that an animal will maintain arbitrary precision about its uncertainty: it is difficult to distinguish between 70\% and 80\% confidence. 

We then define the transition matrix in the discretized belief space by binning the transition matrix, integrating the transition probabilities for the continuous belief space within a given bin. We also allow diffusion between neighboring bins, which reflects additional belief stochasticity. Due to the diffusion, we can only estimate a posterior distribution over the agent's beliefs. More details about the transition probability on belief space can be found in the Supplementary Information.


When a button-press action is taken to open a box, any available reward there is acquired. Afterwards, the animal knows there is no more food available now in the box (since it was either unavailable or consumed) and the belief for that box is reset to zero. 

With the defined transition matrices and reward functions for different actions for the internal model, we can solve for the optimal policy based on the value of different actions. We assume the action of an agent following an optimal strategy is determined rationally according to the value function up to a softmax temperature (Eq \ref{softmax}).

\section{Experiments}
\label{sec:Experiments}
We now apply the learning method for solving an IRC problem (Section \ref{sec:InversePOMDP}) to the foraging task of Section \ref{sec:ForagingTask}. Our goal is to estimate a simulated agent's internal model and belief dynamics from its chosen actions in response to its sensory observations. 

We assume that reward availability at both boxes follows a telegraph process, with appearance rates of $\gamma_1 = 0.15$, $\gamma_2 = 0.1$ and disappearance rate of $\epsilon_1=0.05$, $\epsilon_2=0.04$ respectively, per discrete time step. There are five possible colors in total. Redder colors indicate higher probability that food is available in the box; bluer colors indicate lower probability. The two parameters that influence the color observations are $q_1 = 0.4$, and $q_2 =0.6$. The closer the two values are, the more ambiguous the sensory cue is. Our target agent has made wrong assumptions about these parameters: $\hat{\gamma}_1 = 0.2$, $\hat{\gamma}_2 = 0.15$, $\hat{\epsilon}_1=0.1$, $\hat{\epsilon}_2=0.08$, $\hat{q}_1 = 0.42$, and $\hat{q}_2 =0.66$.

We measure gains and losses in currency of reward, $r \equiv 1$. In those units, the cost (negative reward) of pressing the button is 0.3, and that of traveling is 0.2 (switching between boxes requires two steps, for a total cost of 0.4). We also allow a `grooming' reward for waiting of $r = 0.2$ at the center location. These subjective costs are properties of the agent, and we need to infer them from the agent's behavior.

Our target agent uses a softmax policy (Eq \ref{softmax}) with temperature $\tau = 0.2$. This small temperature enables the agent to follow an approximately optimal policy based on state-action value $Q(b, a)$. 


The actions and sensory evidence (color cues, locations and rewards) obtained by the agent all constitute observations for the experimenter's learning of the agent's internal model. Based on these observations over 5000 time points, we use the EM algorithm to infer the parameters of the internal model that can best explain the behavioral data.

\begin{figure}[h]
   \centering
   \includegraphics[width=5.5in]{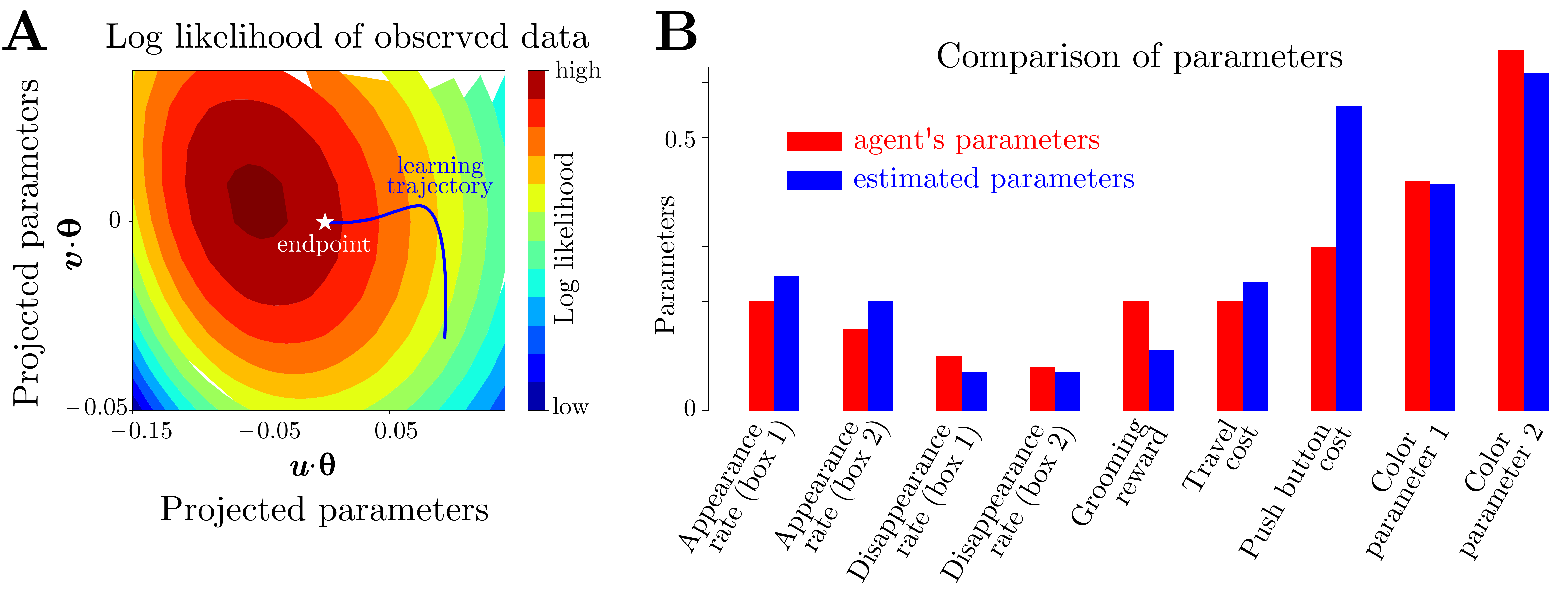}
   \vspace{-10pt}
   \caption{{\bf A}: The estimated parameters converge to the optimal point of the log-likelihood contour. Since the parameter space is high dimensional, we project them onto the first two principal components $\vu,\vv$ of the learning trajectory for $\vtheta$. {\bf B}: Comparison of the true parameters of the agent and the estimated parameters. The agent expects higher appearing rates and lower disappearing rates in both of the boxes. This mismatch is compensated by a higher cost of pushing a button, which accounts for the lower frequency of the agent's attempts to open the box.}
   \label{fig:Para}
\end{figure}

We show the results for inference based on a typical set of data. With the EM algorithm, we found the parameters that have nearly maximal likelihood for the given data (Figure \ref{fig:Para}A). The comparison between the true parameters and the estimated parameters are shown in Figure \ref{fig:Para}B. 

\begin{figure*}[h]
   \centering
   \includegraphics[width=5.5in]{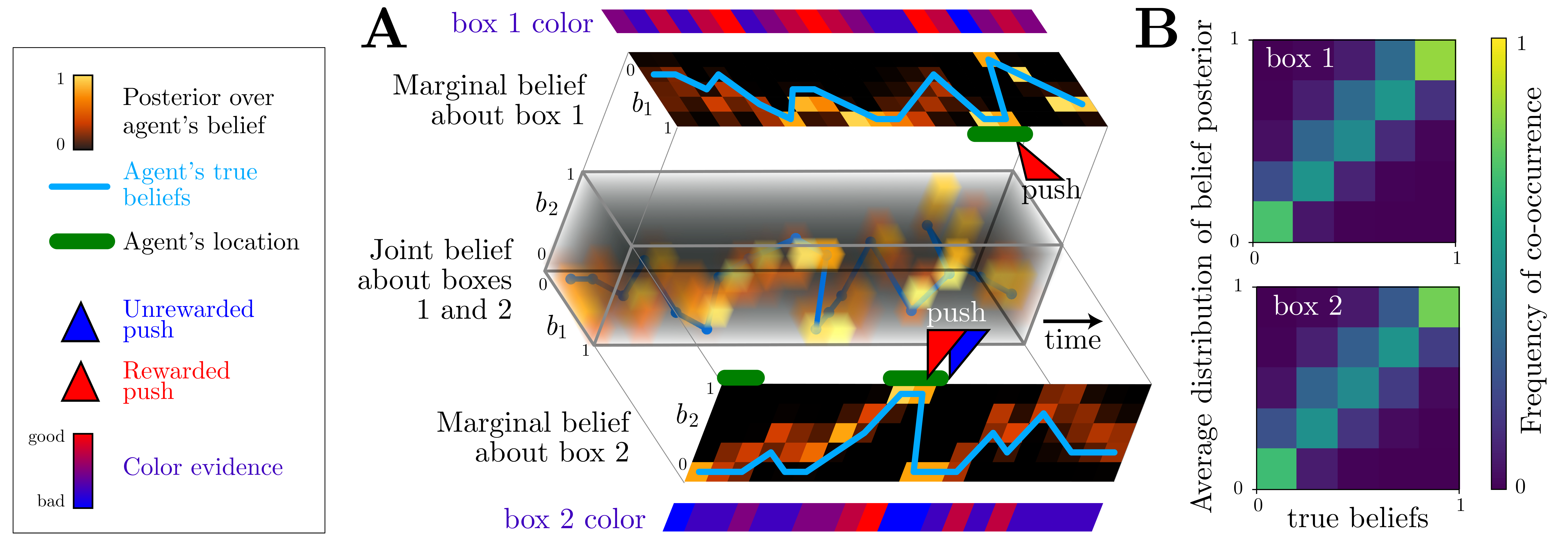}
   \caption{Successful recovery of beliefs by Expectation Maximization. {\bf A}: Estimated and true belief dynamics over latent states. In the 3D plot, we show the agent's time-dependent joint belief about reward availability in the two boxes $(b_{1t},b_{2t})$, as well as our posterior over that joint belief. We also show the marginal belief about box 1 and 2 obtained by projecting the 3d plot along $b_1$ and $b_2$ direction. These estimates are informed by the noisy color data at each box and the times and locations of the agent's actions. The posteriors over beliefs are consistent with the dynamics of the true beliefs. {\bf B}: The averaged posteriors of the estimated beliefs $\hat{b}_t$, $\tfrac{1}{T}\sum_t^T p(\hat{b}_t|a_{0:T},o_{0:T})$, correlates strongly with the agent's true beliefs.}
   \label{fig:posterior}
\end{figure*}

Due to the limited amount of data, there is an expected discrepancy between the true parameters and the estimated parameters. This discrepancy can be reduced with larger amount of data. With the estimated parameters, we are able to infer dynamics of the posterior over the latent states (Figure \ref{fig:posterior}A), which are the beliefs on the two boxes. Note that this is an experimenter's posterior over the agent's subjective posterior. The inferred posteriors is consistent with the true probability of the food availability in each box according to the underlying telegraph process. For each of the two boxes, given the true belief, the distribution of the posterior of beliefs shows strong correlation between the true and estimated belief state, as seen in Figure \ref{fig:posterior}B.

\begin{figure}[h]
   \centering
   \includegraphics[width=5.5in]{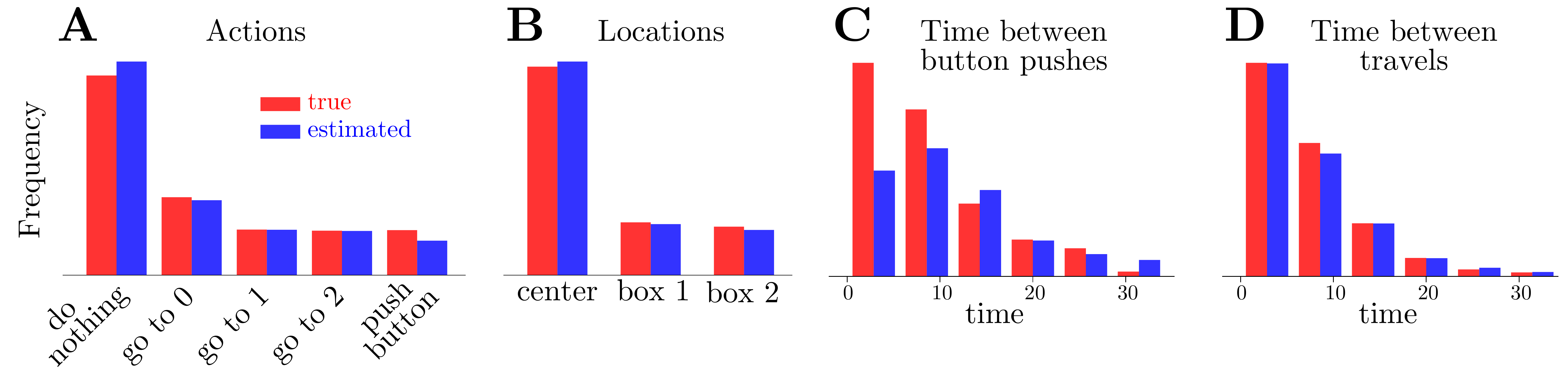}
   \caption{Comparing statistics of behaviors for the actual agent and the inferred agent. {\bf A}: The distribution of actions. {\bf B}: The distribution of time staying at each location. {\bf C}: The distribution of time intervals between two button pressing actions. {\bf D}: The distribution of time intervals between traveling actions.}
   \label{fig:trjectoryComp}
\end{figure}

Since our inferred model parameters differ slightly from the agent's true parameters, we examine how those two internal models differ.  We therefore created another simulated agent using the inferred parameters. 
Figure \ref{fig:trjectoryComp} shows that under softmax near-optimal policies, the two agents choose actions with similar frequencies, occupy the three locations for the same fraction of time, and wait similar amounts of time between pushing buttons or travelling. This demonstrates that our estimated agent's internal model generates behaviors that are consistent with behaviors of the agent from which it learned. 

 \section{Conclusions}
We presented a method to infer the internal model of a rational agent who collects rewards in a task by following a Partially Observable Markov Decision Process. Given that an agent chooses actions in this way, the estimation of its internal model parameters can be formulated as a maximum likelihood problem with latent variables, and the parameters can be inferred using the EM algorithm. When we applied our method to a foraging task, experiments showed that the parameters that best explain the behavior of the agent nicely matched the internal parameters of that agent. The estimated internal model and the true internal model produced similar value functions and behavioral statistics.

Our framework is quite general, and can be applied to still more complex tasks. It can be used to infer false beliefs derived from incorrect or incomplete knowledge of task parameters. It can also be used to infer incorrect {\it structure} within a given model class. For example, it is natural for animals to assume that some aspects of the world, such as reward rates at different locations, are not fixed, even if an experiment actually uses fixed rates \cite{glaze2018bias}. Similarly, an agent may have a superstition that different reward sources are correlated even when they are independent in reality. Given a model class that includes such counterfactual relationships between task variables, our method can test whether an agent holds these incorrect assumptions.

The success of our method on simulated agents suggests our method could be fruitfully applied to experimental data from real animals performing such foraging tasks \cite{sugrue2004matching, odoemene2018visual}. Accurate estimation of dynamic belief states will provide useful targets for interpreting dynamic neural activity patterns, which could help identity the neural substrates of task-relevant thoughts.

\bibliographystyle{unsrt}
\bibliography{References}{}


\end{document}